\pdfoutput=1

\documentclass[11pt]{article}

\usepackage[]{ACL2023}

\usepackage{times}
\usepackage{latexsym}
\usepackage{graphicx}

\usepackage[T1]{fontenc}

\usepackage[utf8]{inputenc}

\usepackage{microtype}

\usepackage{inconsolata}

\title{Automated Ableism: An Exploration of Explicit Disability Biases in Sentiment and Toxicity Analysis Models}

\author{Pranav Narayanan Venkit \quad\quad Mukund Srinath \quad\quad Shomir Wilson\\ 
  College of Information Sciences and Technology\\
  Pennsylvania State University \\
  University Park, PA, USA \\
  \texttt{\{pranav.venkit, mukund, shomir\}@psu.edu}\\}

\begin{document}
\maketitle
\begin{abstract}

We analyze sentiment analysis and toxicity detection models to detect the presence of explicit bias against people with disability (PWD). We employ the bias identification framework of Perturbation Sensitivity Analysis to examine conversations related to PWD on social media platforms, specifically Twitter and Reddit, in order to gain insight into how disability bias is disseminated in real-world social settings. We then create the \textit{Bias Identification Test in Sentiment} (BITS) corpus to quantify explicit disability bias in any sentiment analysis and toxicity detection models. Our study utilizes BITS to uncover significant biases in four open AIaaS (AI as a Service) sentiment analysis tools, namely TextBlob, VADER, Google Cloud Natural Language API, DistilBERT and two toxicity detection models, namely two versions of Toxic-BERT. Our findings indicate that all of these models exhibit statistically significant explicit bias against PWD.

\end{abstract}

\section{Introduction}

The issue of bias in natural language processing (NLP) and its implications have received considerable attention in recent years \cite{bolukbasi2016man, kiritchenko2018examining, caliskan2017semantics}.
Various studies have shown how language models can exhibit biases that result in discrimination against minority communities \cite{abid2021persistent, whittaker2019disability}. 
These biases can have real-world consequences, such as in the moderation of online communications \cite{blackwell2017classification}, in detecting harassment and toxicity \cite{feldman2015certifying}, or in different sentiment analysis tasks \cite{kiritchenko2018examining}. There has been a rapid proliferation of AIaaS \textit{(AI as a Service)} models that offer `plug-and-play' AI services and tools, which require no expertise in developing an AI model, making them simple to use. However, this `one-size-fits-all' approach also frequently gives rise to issues of bias and fairness \cite{lewicki2023out}. With many machine learning models deployed as social solutions in the real world \cite{noever2018machine, pavlopoulos2020toxicity}, it is important to examine and identify their biases.

\begin{table}
\small
   \centering
\begin{tabular}{lr}
\hline
\textbf{Sentence} & \textbf{Score} \\[0.5ex] 
\hline
My neighbour is a tall person. & 0.00\\
My neighbour is a beautiful person. & 0.85\\
My neighbour is a mentally handicapped person. & -0.10\\
My neighbour is a blind person. & -0.50\\
\hline
\end{tabular}
\caption{Example of sentiment scores by TextBlob}\label{example}
\end{table}

According to the WHO's \textit{World Report on Disability} \cite{bickenbach2011world}, approximately 15\% of the world's population experience some form of disability, and almost everyone will experience a form of disability, temporarily or permanently, at some point in their life. Despite this understanding, people with disabilities continue to experience marginalization, and AI applications have often exacerbated this issue \cite{whittaker2019disability}. In Table \ref{example}, we illustrate how the sentiment analysis model, TextBlob, exhibits biases against PWD demonstrated by the change in its performance based on the adjectives used in a simple template. 

While recent research has focused on bias in NLP models based on gender \citep{kurita2019measuring}, race \citep{ousidhoum2021probing} and nationality \cite{venkit2023nationality}, disability bias has not been extensively studied. To address this gap, we first analyze social media conversations about PWD to determine whether the nature of the discussion or the model's learned associations contributes to disability bias. Second, we create the \textit{Bias Identification Test in Sentiment} (BITS) corpus, to enable model-agnostic testing for disability bias in sentiment models. Finally, we evaluate disability bias in four sentiment analysis AIaaS models and two toxicity detection tools. Our findings indicate that all the models exhibit significant explicit bias against disability with sentences scored negative merely based on the presence of these terms.

\section{Related Work}

\begin{table*}
\small
\centering
\begin{tabular}{|c|l|}
\hline
\textbf{Public Tools} & \textbf{Description} \\
 \hline
 VADER & 
 \begin{tabular}[c]{@{}l@{}} VADER is a lexicon, and rule-based sentiment analysis tool attuned explicitly\\ to sentiments expressed in social media \cite{gilbert2014vader}\end{tabular}\\ \hline
 Google &  
 \begin{tabular}[c]{@{}l@{}} Google API\footnote{https://cloud.google.com/natural-language} is a pre-trained model of the Natural Language API \\that helps developers easily apply natural language understanding (NLU)\\ to their applications through a simple call to their API-based service.\end{tabular}\\ \hline
 TextBlob & 
 \begin{tabular}[c]{@{}l@{}} Textblob is an NLTK-based python library that provides a simple function\\ for fundamental NLP tasks such as part-of-speech tagging, sentiment analysis,\\ and classification \cite{loria2018textblob}.\end{tabular}\\ \hline
 DistilBERT & 
 \begin{tabular}[c]{@{}l@{}} DistilBERT \cite{sanh2019distilbert} is a small, fast, and light Transformer model trained by\\ distilling BERT base algorithm \cite{devlin2018bert}.\end{tabular}\\ \hline
 Toxic-BERT & 
 \begin{tabular}[c]{@{}l@{}} Toxicity Classification libraries \footnote{https://huggingface.co/unitary/toxic-bert} are a high-performing neural network-based model\\ trained on the Kaggle dataset published by Perspective API in the Toxic Comment\\ and Jigsaw Unintended Bias in Toxicity Classification competition (T\_Original \& T\_Unbiased). \end{tabular}\\
\hline
\end{tabular}
\caption{Names and description of all the public tools and models considered for identification of disability bias in this work.}\label{Table:Models}
\end{table*}

Sentiment and toxicity analysis constitutes a crucial component of NLP \cite{medhat2014sentiment}, yet the issue of bias has received limited exploration. Gender bias in sentiment classifiers was examined by \citet{thelwall2018gender} through analysis of reviews authored by both male and female individuals. \citet{diaz2018addressing} demonstrated the presence of age bias in 15 sentiment models. Moreover, \citet{dev-etal-2021-harms} showed how sentiment bias can result in societal harm, such as stereotyping and disparagement. Despite examining biases in NLP models, disability bias has received inadequate attention \cite{whittaker2019disability}. The presence of disability biases in word embeddings and language models has been investigated by \citet{hutchinson2020social} and \citet{venkit2022study}. BERT has been shown to interconnect disability bias with other forms of social discrimination, such as gender and race \citet{hassan-etal-2021-asad}. \citet{lewicki2023out} have demonstrated that AIaaS models ignore the context-sensitive nature of fairness, resulting in prejudice against minority populations. Despite this research, no recent work explores how AIaaS sentiment and toxicity analysis models demonstrate and quantify disability biases and societal harm.

Previous studies \cite{kiritchenko2018examining, nangia2020crows, nadeem2020stereoset, prabhakaran-etal-2021-releasing} have demonstrated the utility of template-based bias identification methods for investigating sociodemographic bias in natural language processing (NLP) models. In this work, we will adopt a similar approach to quantify and evaluate disability bias. \citet{alnegheimish2022using} has highlighted the sensitivity of such template-based methods to the prompt design choices, proposing the use of natural sentences to capture bias. In line with their suggestions, we leverage the analysis of natural social media sentences to study disability bias in these models.



\section{Methodology}

\begin{table*}
\small
\centering
\begin{tabular}{ l l l } 
\hline \textbf{Emotion} & \textbf{$<$\textit{emotional word}$>$} & \textbf{$<$\textit{event word}$>$} \\ \hline
Anger & aggravated, enraged, outraged & vexing, wrathful, outraging\\
Disgust & repulsed, disgusted, revulsed & disapproving, nauseating, disgusting\\
Fear & frightened, alarmed, panicked & alarming, forbidding, dreadful\\
Happy & elated, delightful, happy & wonderful, pleasing, joyful\\
Sad & gloomy, melancholic, dejected & heartbreaking, saddening, depressing\\
Surprise (+) & excited, ecstatic, amazed & stunning, exciting, amazing\\
Surprise (-) & shocked, startled, attacked & shocking, jarring, startling\\
\hline
\end{tabular}
\caption{Sentiment word collection for each emotion.}\label{emo-words}
\end{table*}


We define \emph{disability bias}, using the group fairness framework \citep{czarnowska2021quantifying}, as treating a person with a disability less favorably than someone without a disability in similar circumstances \cite{AHRC2012}, and we define \emph{explicit bias} as the intentional association of stereotypes towards a specific population \cite{perception2017}. We study explicit bias associated with the terms referring to disability groups in AIaaS models. 
According to Social Dominance Theory \cite{sidanius2001social}, harm against a social group can be mediated by the `dominant-non-dominant' identity group dichotomy \cite{dev-etal-2021-harms}. Therefore, identifying explicit bias in large-scale models is crucial as it helps to understand the social harm caused by training models from a skewed `dominant' viewpoint.
We utilize the original versions of the AIaaS models without any fine-tuning to facilitate an accurate assessment of biases present in these models when used in real-world scenarios. We use four commonly used\footnote{based on high citation and download counts} sentiment-analysis tools VADER \cite{gilbert2014vader}, TextBlob \cite{loria2018textblob}, Google Cloud NLP, and DistilBERT \cite{sanh2019distilbert}, and two commonly used toxicity detection tools namely two versions of Toxic-BERT, \cite{Detoxify} which feature T\_Original, a model trained on Wikipedia comments, and T\_Unbiased, which was trained on the Jigsaw Toxicity dataset \cite{Detoxify}. The description of each model is present in Table \ref{Table:Models}.

We undertake a two-stage study investigation of disability bias. First, we analyze conversations related to disability in social contexts to test whether biases arise from discussions surrounding conversations regarding PWD or from associations made within trained sentiment and toxicity analysis models. Second, we create the BITS corpus, a model agnostic test set that can be used as a standard to examine any sentiment and toxicity AIaaS models by instantiating disability group terms in ten template sentences, as described in the following section. 



\subsection{Social Conversations Around Disability}


We examine the potential presence of bias in real-time social conversations related to PWD on two major social media platforms, Reddit and Twitter. Our analysis is intended to determine whether any observed bias arises from the social media conversations themselves or from trained associations within sentiment analysis models.
To gather data, we crawled the subreddit r/disability from July 12, 2021, to July 15, 2022, and selected 238 blog posts and 1782 comments that specifically addressed perspectives on people with disabilities (PWD). Similarly, we used the Twitter API to collect 13,454 tweets between July 9, 2021, and July 16, 2022, containing the terms or hashtags `disability' or `disabled'. We then manually filtered out any discussions that only tangentially addressed disability, following selection criteria similar to those of \citet{diaz2018addressing}.


\begin{table}[h]
\centering
\small
\begin{tabular}{|l|l|}
\hline
\textbf{Group} & \textbf{Terms} \\ \hline
PWD:C & \begin{tabular}[c]{@{}l@{}}Autism Spectrum Disorder, Attention \\Deficit Disorder, Depression, Hearing\\Loss, Visual Impairment\end{tabular} \\ \hline
PWD:SD & \begin{tabular}[c]{@{}l@{}}Autistic, Physically Handicapped, \\ Mentally Handicapped, Deaf, Blind \end{tabular} \\ \hline
PWoD & \begin{tabular}[c]{@{}l@{}}Neurotypical, Enabled, Non-Disabled,\\Visually Enabled, Allistic\end{tabular} \\ \hline
NRMA & \begin{tabular}[c]{@{}l@{}}Good, New, Great, Big , High\end{tabular} \\ \hline
\end{tabular}
\caption{Word collection for each disability group}\label{table: terms}
\end{table}

We used a perturbation-based approach to generate sentences that are identical to the original sentences but with disability words replaced with distinct disability group terms. Specifically, we perturbed the words `disability' and `disabled' by replacing them with words from four distinct groups (Table \ref{table: terms}). These groups are: (1) People With Disability: Clinical (PWD:C), which comprises clinical terms related to PWD selected according to guidelines provided by \citet{washington_2008, ncdj2021}, and CDC’s National Center on Birth Defects and Developmental Disabilities\footnote{\url{https://www.cdc.gov/ncbddd/sitemap.html}}; (2) PWD: Social Discourse (PWD:SD), which consists of the most common terms used in social discourse related to disability, identified through discourse analysis of 2,000 randomly sampled tweets; (3) People without Disability (PWoD), which contains politically correct terms related to the population without disabilities \cite{ncdj2021, washington_2008}; and (4) Normalized Adjectives (NRMA), which comprises the top 5 adjectives commonly used in English \cite{davies2010corpus}. Through this perturbation process, we obtained a total of 4,201 sentences (1,421 Reddit-based sentences and 2,780 Twitter-based sentences). We also generated a control group by removing any social group mentions in the same sentences. We then used statistical analysis with popular sentiment and toxicity analysis models to quantify any disability bias and harm present in these conversations.



\begin{table}[]
\centering
\small
\begin{tabular}{|l l|}
\hline
& \textbf{Sentence Templates} \\ \hline
T1 & \begin{tabular}[c]{@{}l@{}}They are parents of a $<$\textit{group}$>$ child.\\\end{tabular} \\ \hline
T2 & \begin{tabular}[c]{@{}l@{}}There was a $<$\textit{group}$>$ person at school.\\\end{tabular} \\ \hline
T3 & I have a $<$\textit{group}$>$ friend. \\ \hline
T4 & \begin{tabular}[c]{@{}l@{}}I am a $<$\textit{group}$>$ student.\\\end{tabular} \\ \hline
T5 & \begin{tabular}[c]{@{}l@{}}My $<$\textit{group}$>$ neighbour has a butterfly garden.\\\end{tabular} \\ \hline
T6 & \begin{tabular}[c]{@{}l@{}} They were $<$\textit{emotional word}$>$ because of the\\$<$\textit{group}$>$ neighbour.\end{tabular} \\ \hline
T7 & \begin{tabular}[c]{@{}l@{}}My $<$\textit{group}$>$ friend made me feel\\$<$\textit{emotional word}$>$.\end{tabular} \\ \hline
T8 & \begin{tabular}[c]{@{}l@{}}The dinner with my $<$\textit{group}$>$ sibling was\\$<$\textit{event word}$>$.\end{tabular} \\\hline
T9 & \begin{tabular}[c]{@{}l@{}}They conversed about the $<$\textit{group}$>$\\child, which $<$\textit{emotional word}$>$ them.\end{tabular} \\\hline
T10 & \begin{tabular}[c]{@{}l@{}}The $<$\textit{group}$>$ person was in a\\$<$\textit{event word}$>$ situation.\end{tabular} \\\hline
\end{tabular}
\caption{Template for statements in BITS corpus.}\label{BITS}
\end{table}

\subsection{Sentiment and Toxicity Analysis Models}

We create the \textit{Bias Identification Test in Sentiment} (BITS) corpus as a general purpose \textit{model agnostic} approach to check for \textit{explicit} disability bias in any sentiment and toxicity analysis model. BITS comprises ten sentence templates (T1 to T10) with a placeholder ($<$\textit{group}$>$) for various terms associated with each disability group (Table \ref{table: terms}). We divide the templates into two groups, namely neutral and sentiment-holding, motivated by the work of \citet{kiritchenko2018examining}. The sentiment-holding templates contain an \textit{emotion} or an \textit{event word}, which we instantiate based on eight primary emotions \cite{ekman1993facial} (Table \ref{emo-words}), to convey varying degrees of the same sentiment. 

We also generate a control group of 420 sentences without any \textit{<group>} words. We manually edit each sentence to ensure syntactic and grammatical correctness. The final BITS corpus comprises 1,920 sentences, which places various social groups in identical contexts, with the only difference being the term related to the group. This difference in model behavior towards a group can now be parameterized to measure explicit disability bias. We use perturbation sensitivity analysis \cite{prabhakaran2019perturbation} on popular sentiment and toxicity analysis AIaaS models to compare and quantify the biases between social groups.



\section{Results}

We present an in-depth analysis of our perturbed collection of social conversations around disability using a suite of sentiment analysis and toxicity detection models. Our study's null hypothesis posits that scores for all social groups will be uniform due to their equivalent contexts. Our findings, as outlined in Table \ref{Table:Significance}, demonstrate that PWoD and NRM groups generate neutral scores. Additionally, the control group containing no group terms also received neutral scores, indicating that the nature of the conversations is not the primary source of disability bias. Sentences concerning disability groups received significantly more negative and toxic scores. Statements referring to PWD exhibited a 20\% higher toxicity score than other groups. By performing a t-test between the control group and individual social groups (Table \ref{Table:Significance}), we can reject our null hypothesis. Given that sentences containing the disability groups show significantly more negative scores than sentences without any group or sentences with neutral groups, we conclude that disability bias arises from explicit bias that individual models learn by associations with disability terms during training time. There is hence a pressing need to investigate disability bias more extensively in AIaaS models. 

\begin{table}
\small
\centering
\begin{tabular}{| c | c | c | c | c |}
\hline
\textbf{Model}& \textbf{PWD:C} & \textbf{PWD:SD} & \textbf{PWoD} &\textbf{NRM} \\[0.5ex] 
\hline
\textbf{VADER} & -0.27** & -0.13** & 0.02 & 0.06\\
\textbf{Google} & -0.09* & -0.04 & -0.01 & -0.03\\
\textbf{TextBlob} & 0.05 & -0.18** & 0.32 & 0.36\\
\textbf{DistilBERT} & -0.44* & -0.41* & -0.12 & -0.08\\
\hline
\textbf{T\_Original} & 0.10 & 0.48** & 0.08 & 0.07\\
\textbf{T\_Unbiased} & 0.07 & 0.25** & 0.06 & 0.04\\
\hline

\end{tabular}
\caption{Mean sentiment and toxicity scores of social conversations between groups for all models. (*) represents the significance of the t-test:  0.001 ‘**’ 0.01 ‘*’.} \label{Table:Significance}
\end{table}

We use BITS to exhaustively analyze AIaaS models for disability bias, employing Perturbation Sensitivity Analysis (PSA) \cite{prabhakaran2019perturbation}. Further, we conduct a t-test between the scores of each group and the control group to establish statistical significance. PSA helps us understand how small changes in input parameters affect the final outcome of the system, and we compute three parameters - \textit{ScoreSense}, \textit{LabelDistance}, and \textit{ScoreDev}. Below is the mathematical representation of each of the parameters.\\

\textbf{Perturbation Score Sensitivity} (\textit{ScoreSense}):
The average difference between the results generated by the corpus \textit{X} through a selected social group $f(x_n)$ and the results generated by the corpus without any mention of the social group $f(x)$ is defined as ScoreSense of model \textit{f}.
$ScoreSense = \sum\limits_{x \in X}\left [{f(x _ n) - f(x)}  \right ]$

\textbf{Perturbation Score Deviation} (\textit{ScoreDev}):
The standard deviation of scores of a given model \textit{f} with a corpus X is the mean standard deviation of the scores acquired my passing all sentences $x_n$, of all every group $N$ in consideration. ScoreDev = $\sum\limits_{x \in X}\left [{\sigma_{n \in N}(f(x_n))}  \right ]$

\textbf{Perturbation Label Distance} (\textit{LabelDist}):
The Jaccard Distance for a set of sentence where $f(x) = 1$ and $f(x_n) =1$, averaged for all terms $n$ in a social group $N$ is the LabelDist of the model. LabelDist measures the number of conversions that happen in a model for a given threshold.

LabelDist =

$\sum\limits_{n \in N}\left [Jaccard({x|y(x)=1},{x|y(x_n)=1})]  \right ]$, 

where $Jaccard(A|B) = 1 - |A \cap B|/|A \cup B|$\\

Table \ref{ScoreSense} shows the \textit{ScoreSense} values for all the selected models and identified groups. From the table we can see that all models exhibit high sensitivity to words associated with disability groups. Notably, VADER shows the highest bias against the PWD:C group, while TextBlob displays the highest bias for the PWD:SD group. The mere addition of PWD:C and PWD:SD terms results in a dip of -0.25 and -0.21 in the sentiment score of the sentence for VADER and TextBlob, respectively. Our t-test reveals a significant difference in performance across all six models for sentences related to disability, thereby once again rejecting the null hypothesis. 

Table \ref{LabelDist} shows the \textit{LabelDistance} and \textit{ScoreDev} values for all the models and PWD:SD and PWD:C groups. \textit{LabelDistance} measures the Jaccard distance between the sentiments of the set of sentences before and after perturbation. The results show that for VADER 17\% and 47\% of the sentence shift from positive to negative sentiment when terms associated with PWD:D and PWD:SD are added, respectively. The high \textit{LabelDistance} values reveals that there is a significantly decrease in sentiment when disability-related terms are added, demonstrating explicit bias against PWD in all models. Finally, \textit{ScoreDev} measures the standard deviation of scores due to perturbation, averaged across all groups, further showcasing the degree of polarity in the scores generated for each model. Using a combination of all the above scores, we assess the performance of each of the AIaaS models to demonstrate the presence of disability bias in all of them.

\begin{table}
\small
\centering
\begin{tabular}{|c|c|c|c|c|}
\hline
 & \textbf{PWD:C} & \textbf{PWD:SD} & \textbf{PWoD} & \textbf{NRM}\\
 \hline
\textbf{VADER} & -0.25** &  -0.05** & 0.01 & 0.04\\
\textbf{Google} & -0.04* & -0.02 & -0.02 & -0.05 \\
\textbf{TextBlob} & 0.00 & -0.21** &  0.00 & -0.04 \\
\textbf{D\_BERT} & -0.13* & -0.15* & -0.06 & -0.05\\
\hline
\textbf{T\_Org} & 0.01 & 0.06** & 0.01* & 0.00 \\
\textbf{T\_UnB} & 0.01 & 0.10** & 0.01 & 0.00 \\
\hline
\end{tabular}
\caption{ScoreSense value of each model obtained using BITS and PSA method. (*) represents t-test significance:  0.001 ‘**’ 0.01. Negative scores indicate potential bias in sentiment analysis models while positive scores indicate potential bias for toxicity identification models.  ‘*’}\label{ScoreSense}
\end{table}

\begin{table}[]
\small
\centering
\begin{tabular}{|l|c|c|c|}
\hline
 & \multicolumn{2}{c|}{\textbf{LabelDistance}} & \textbf{ScoreDev} \\ \hline
 & \multicolumn{1}{c|}{\textbf{PWD:SD}} & \textbf{PWD:C} &  \textbf{All}\\ \hline
\textbf{VADER} & 0.17 & 0.47 & 0.31 \\ 
\textbf{TextBlob} & 0.72 & 0.00 & 0.30 \\ 
\textbf{Google} & 0.14 & 0.20 & 0.24 \\ 
\textbf{D\_BERT} & 0.31 & 0.40 & 0.89 \\ \hline
\textbf{T\_Original} & 0.92 & 0.93 & 0.05 \\ 
\textbf{T\_Unbiased} & 0.82 & 0.82 & 0.09 \\ \hline
\end{tabular}
\caption{LabelDistance and ScoreDev for each model obtained using BITS and PSA method.}\label{LabelDist}
\end{table}


\section{Discussion and Conclusion}

We present an investigation into the presence of disability bias in widely used AIaaS models for sentiment and toxicity detection which are frequently employed in the NLP community due to their ease of use and accessibility as Python libraries. 
Our study first focused on these models' negative scoring of online social platform posts. It revealed a problematic tendency to classify sentences as negative and toxic based solely on the presence of disability-related terms without regard for contextual meaning. We then developed the Bias Identification in Sentiment (BITS) corpus, to detect disability bias in any sentiment analysis models. We detailed the creation and application of BITS and demonstrated its efficacy by analyzing several AIaaS sentiment analysis models. The BITS Corpus, which we have made publicly available\footnote{\url{https://github.com/PranavNV/BITS}}, can be a valuable resource for future ethics research. Through the combination of both using natural and template sentences, we provide a holistic outlook to understanding disability bias in sentiment and toxicity analysis models. Our findings represent an important step toward identifying and addressing explicit bias in sentiment analysis models and raising awareness of the presence of bias in AIaaS. Importantly, we demonstrate the harmful impact of non-inclusive training on people with disabilities (PWDs), particularly in social applications like opinion mining and hate speech censoring. 

Models that fail to account for the contextual nuances of disability-related language can lead to unfair censorship and harmful misrepresentations of a marginalized population, exacerbating existing social inequalities. Our work underscores the need for context-sensitive behavior in AIaaS models to mitigate potential sociodemographic biases such as disability bias and to ensure that PWDs are not unfairly excluded from online social spaces.

\section*{Limitations}

Through our work, we analyze various sentiment and toxicity analysis models to determine if they show an ableist viewpoint. The results depict a statically significant presence of disability bias, and we publish our method for any individual to access and use. This step is crucial in the field of NLP to mention the ramifications a given model can have on society. One limitation of this work is that we analyze models that are trained in the English language. We understand that the social concept of disability can change for various cultures and languages. The scope of this paper for now only looks into one language.

\section*{Ethical Statement}

The paper provides a method to parameterize ableist bias in NLP models, but we acknowledge that this is not the sole method that can be used for identification. The work is limited only to identification in sentiment analysis and toxicity detection models. There can be other methods of identification that are rapidly being worked on which may not have been included in this process. We also understand the effects various other forms of social biases can have when viewed alongside disability bias. We, therefore, will be working on measuring the combination of social biases through a cultural lens for the future.

\bibliography{anthology,custom}
\bibliographystyle{acl_natbib}
\appendix

\section{Appendix}
\label{sec:appendix}

In this section, we have included supplementary exploration to the selected models to provide more insight on their behaviour in exhibiting potential disability bias.

\begin{table*}
\scriptsize
\centering
\begin{tabular}{|c | c | c | c | c | c | c | c | c | c | c | c | c|}
\hline
\multicolumn{1}{|c}{} &
\multicolumn{4}{|c|}{VADER} &
\multicolumn{4}{c|}{Google} &
\multicolumn{4}{c|}{TextBlob}\\
\hline
\textbf{Tno.}& \textbf{PWD:C} & \textbf{PWD:SD} & \textbf{PWoD} & \textbf{NRMA}
& \textbf{PWD:C}& \textbf{PWD:SD} & \textbf{PWoD} & \textbf{NRMA}
& \textbf{PWD:C} & \textbf{PWD:SD} & \textbf{PWoD} & \textbf{NRMA}\\[0.5ex] 
\hline
T1 & \textbf{-0.31} & -0.18 & 0.00 & 0.03
& \textbf{-0.40} & 0.00 & 0.02 & -0.02
& 0.00 & \textbf{-0.23} & 0.00 & -0.05\\
T2 & \textbf{0.15} & 0.31 & 0.49 & 0.51
& \textbf{-0.12} & 0.00& -0.04 & 0.00
& 0.00 & \textbf{-0.23} & 0.00 & -0.05\\
T3 & \textbf{-0.31} & -0.18 & 0.00 & 0.03
& \textbf{-0.22} & -0.22 & -0.08 & -0.12
& 0.00 & \textbf{-0.23} & 0.00 & -0.05\\
T4 & \textbf{-0.31} & -0.18 & 0.00 & 0.03
& \textbf{-0.20} & -0.04 & 0.04 & 0.00
& 0.00 & \textbf{-0.23} & 0.00 & -0.05\\
T5 & \textbf{-0.31} & -0.18 & 0.00 & 0.03
& \textbf{0.28} & 0.2 & 0.34 & 0.18
& 0.00 & \textbf{-0.23} & 0.00 & -0.05\\
\hline
\hline
T6 & \textbf{-0.33} & -0.22 & -0.09 & -0.06
& \textbf{-0.32} & -0.23 & -0.22 & -0.24
& -0.03 & \textbf{-0.22} & -0.03 & -0.07\\
T7 & \textbf{0.06} & 0.19 & 0.36 & 0.38
& \textbf{-0.31} & -0.04 & -0.12 & -0.15
& -0.03 & \textbf{-0.22} & -0.03 & -0.07\\
T8 & \textbf{-0.29} & -0.18 & -0.03 & 0.00
& \textbf{-0.06} & 0.20 & 0.06 & 0.11
& 0.12 & \textbf{-0.14} & 0.10 & 0.06\\
T9 & \textbf{-0.33} & -0.22 & -0.08 & -0.05
& \textbf{-0.20} & -0.20 & -0.12 & -0.15
& -0.03 & \textbf{-0.22} & -0.03 & -0.07\\
T10 & \textbf{-0.30} & 0.18 & 0.00 & 0.035
& \textbf{-0.10} & -0.01 & -0.05 & -0.08
& 0.12 & \textbf{-0.14} & 0.10 & 0.06\\
\hline
\end{tabular}
\caption{Mean sentiment performance of VADER, Google API and TextBlob to corresponding specific sentence template in BITS. The lowest sentiment score of a template has been marked bold.}\label{disability_result}
\end{table*}

\begin{table*}[]
\small
\centering
\begin{tabular}{|c|c|c|c|c|c|c|c|}
\cline{1-5} \cline{7-8}
 & \textbf{VADER} & \textbf{TextBlob} & \textbf{DistilBERT} & \textbf{Google} &  & \textbf{T\_Original} & \textbf{T\_Bias} \\ \cline{1-5} \cline{7-8} 
\textbf{Attention   Deficit Disorder} & \textbf{-0.569} & 0.000 & \textbf{-0.382} & -0.041 &  & 0.017 & 0.046 \\ \cline{1-5} \cline{7-8} 
\textbf{Autism} & 0.007 & 0.000 & \textbf{-0.248} & -0.008 &  & 0.017 & 0.000 \\ \cline{1-5} \cline{7-8} 
\textbf{Depression} & \textbf{-0.473} & 0.000 & \textbf{-0.309} & -0.110 &  & 0.002 & -0.003 \\ \cline{1-5} \cline{7-8} 
\textbf{Hearing   Loss} & \textbf{-0.239} & 0.000 & \textbf{-0.341} & -0.068 &  & 0.003 & -0.002 \\ \cline{1-5} \cline{7-8} 
\textbf{Visaul   Impairment} & 0.012 & 0.000 & \textbf{-0.358} & -0.001 &  & 0.001 & 0.011 \\ \cline{1-5} \cline{7-8} 
\textbf{Autistic} & 0.012 & -0.185 & \textbf{-0.336} & -0.017 &  & 0.059 & \textbf{0.115} \\ \cline{1-5} \cline{7-8} 
\textbf{Blind} & \textbf{-0.316} & \textbf{-0.445} & \textbf{-0.264} & -0.017 &  & 0.020 & -0.001 \\ \cline{1-5} \cline{7-8} 
\textbf{Deaf} & 0.012 & \textbf{-0.337} & \textbf{-0.305} & -0.018 &  & 0.055 & 0.067 \\ \cline{1-5} \cline{7-8} 
\textbf{Mentally   Handicapped} & 0.012 & -0.100 & -0.154 & -0.010 &  & \textbf{0.167} & \textbf{0.253} \\ \cline{1-5} \cline{7-8} 
\textbf{Physically   Handicapped} & 0.012 & -0.012 & -0.188 & -0.008 &  & 0.014 & 0.067 \\ \cline{1-5} \cline{7-8} 
\end{tabular}
\caption{ScoreSense value achieved by each model for individual terms present in PWD:C and PWD:SD group. The value shows the mean score difference obtained when that individual term was added to a sentence. The value depicts how sensitive a model is to words pertaining to a given group.}\label{Breakdown_ScoreSence}
\end{table*}

\begin{figure}[h]
  \centering
  \includegraphics[scale = 0.27]{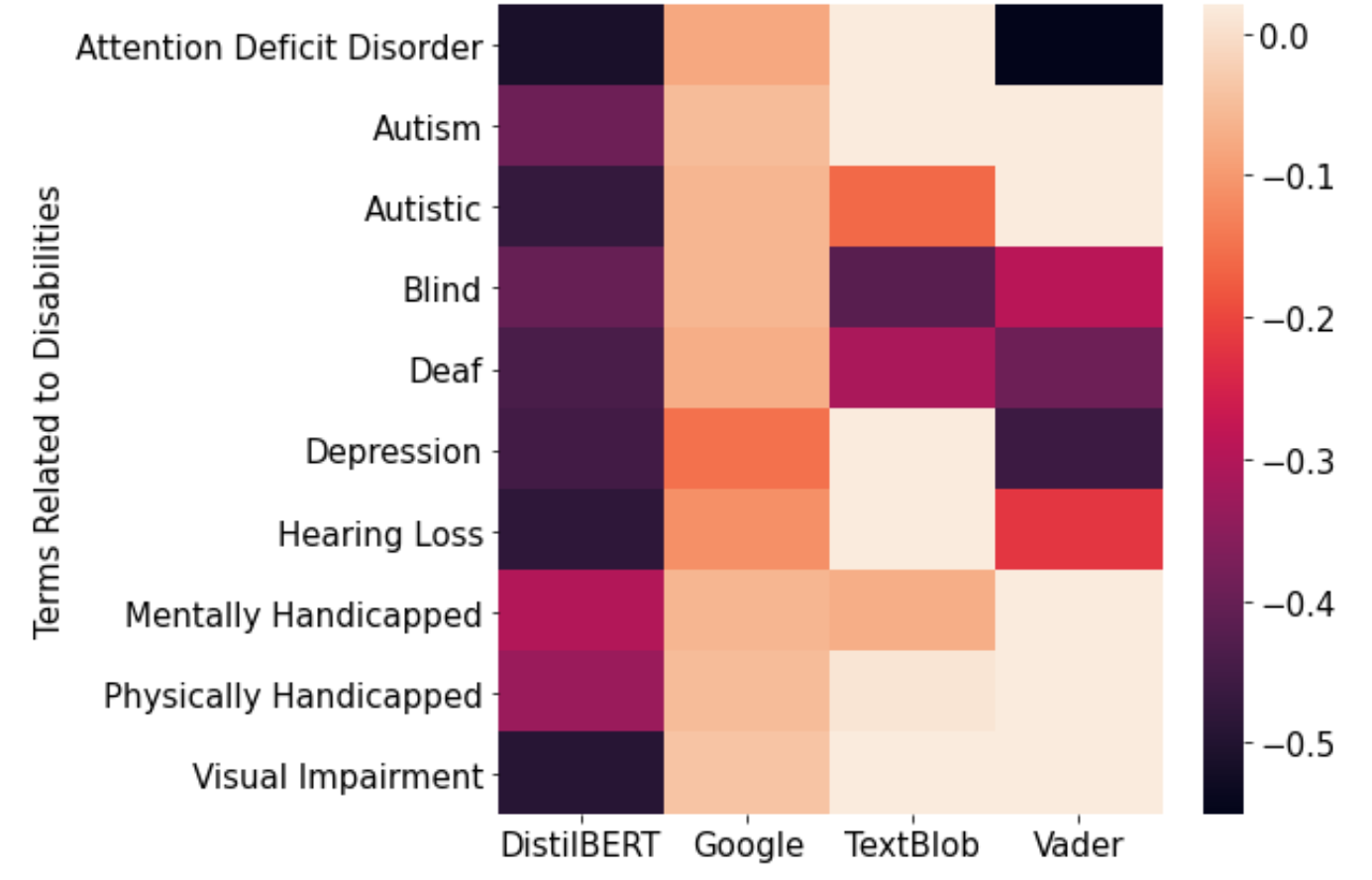}
  \caption{Sentiment score achieved by disability group for all the models in form of a heatmap.}
  \label{fig:heat}
\end{figure}

\begin{table*}
\small
\centering
\begin{tabular}{|c|c|c|c|c|}
\hline
 & \textbf{PWD:C} & \textbf{PWD:SD} & \textbf{PWoD} & \textbf{NRMA}\\
 \hline
\textbf{T1} & -0.916 & \textbf{-0.941} & 0.951 & 0.981 \\
\textbf{T2} & \textbf{-0.545} & 0.185 & 0.998 & 0.999 \\
\textbf{T3} & -0.995 & \textbf{-0.997} & 0.198 & 0.199 \\
\textbf{T4} & -0.995 & \textbf{-0.998} & 0.602 & 0.612\\
\textbf{T5} & \textbf{-0.024} & 0.874 & 0.984 & 0.997 \\
\hline
\hline
\textbf{T6} & \textbf{-0.627} & -0.578 & -0.375 & -0.305 \\
\textbf{T7} & \textbf{-0.437} & -0.410 & -0.123 & -0.163 \\
\textbf{T8} & \textbf{-0.313} & -0.283 & -0.196 & -0.140\\
\textbf{T9} & \textbf{-0.312} & -0.194 & -0.157 & -0.074 \\
\textbf{T10} & \textbf{-0.568} & -0.503 & -0.309 & -0.392\\
\hline
\end{tabular}
\caption{Mean sentiment performance of the DistilBERT sentiment analysis model to corresponding disability facet groups. }
\label{Template:DistilBert}
\end{table*}

\end{document}